# FAST LEXICALLY CONSTRAINED VITERBI ALGORITHM (FLCVA): SIMULTANEOUS OPTIMIZATION OF SPEED AND MEMORY

*Alain Lifchitz*[*]  *Frederic Maire*  *Dominique Revuz*

Laboratoire d'Informatique de Paris 6  
Université P. & M. Curie & CNRS (UMR 7606)  
8, rue du Capitaine Scott  
75015 Paris, France  
alain.lifchitz@lip6.fr

School of S.E.D.C.  
Faculty of Information Technology  
2 George Street, GPO Box 2434  
Brisbane Q4001, Australia  
f.maire@qut.edu.au

Laboratoire d'Informatique  
Institut Gaspard Monge  
bât. Copernic, 5, boulevard Descartes  
77454 Marne-la-vallée Cedex 2, France  
dominique.revuz@univ-mlv.fr

**ABSTRACT**

*Lexical constraints on the input of speech and on-line handwriting systems improve the performance of such systems. A significant gain in speed can be achieved by integrating in a digraph structure the different Hidden Markov Models (HMM) corresponding to the words of the relevant lexicon. This integration avoids redundant computations by sharing intermediate results between HMM's corresponding to different words of the lexicon. In this paper, we introduce a token passing method to perform simultaneously the computation of the a posteriori probabilities of all the words of the lexicon. The coding scheme that we introduce for the tokens is optimal in the information theory sense. The tokens use the minimum possible number of bits. Overall, we optimize simultaneously the execution speed and the memory requirement of the recognition systems.*

## 1. INTRODUCTION

A number of pattern recognition problems like hand gesture recognition, on-line and off-line *Hand Writing Recognition* (HWR) and *Automatic Speech Recognition* (ASR) can be solved by performing an elastic matching between an input pattern and a set of prototype patterns. In all these applications, the a posteriori probabilities of a number of different words are computed given a sequence of frames (feature vectors). These a posteriori probabilities are computed by running *Viterbi Algorithm* (VA) [14, 3] on the *Hidden Markov Models* (HMM) corresponding to the different words [10].

Most cursive HWR and ASR systems use a lexical constraint to help improve the recognition performance. Traditionally, the lexicon is stored in a *trie* [4]. This approach has been extended with solutions based on a more compact data structure, the *Directed Acyclic Word Graph* (DAWG) [5, 13, 6]. The non-deterministic node-automata we use to represent the lexicons can be significantly more compact than their deterministic counterparts [7]. **Figure 2** shows a non-deterministic node-automaton generating the same language as the trie of **Figure 1**.

Node-automata are better at HMM factorization because in a node-automaton the processing is done in the nodes and the routing is done with the arcs, whereas with traditional automata (that we call arc-automata), these two tasks are not separated. In a nutshell, the nodes of our automata encapsulate HMM corresponding to letters. The resulting super-structure is called a *lexicon-HMM*.

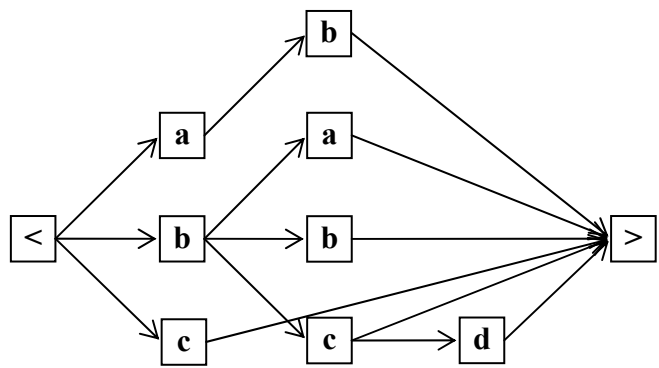

**Figure 1**. A toy example showing the node-automaton associated to a trie. This node-automaton represents the six word lexicon {'ab', 'ba', 'bb', 'bc', 'bcd', 'c'} of 12 letters. This automaton is a trie with an added common sink for all the leaves. The automaton contains 10 nodes.

After recalling the basics of Viterbi algorithm and describing some improvements in Section 2, we present an optimal token tagging scheme for Viterbi algorithm in Section 3.

---

[*] *Correspondence to* : A. Lifchitz

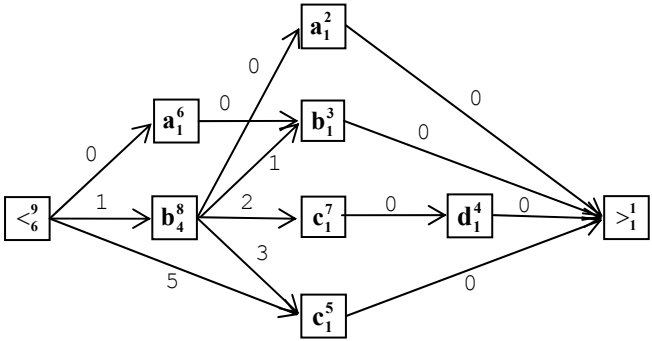

**Figure 2**. The six word toy lexicon of **Figure** 1 as DAWG node-automaton. The automaton contains 9 nodes. Nodes are labeled with a reverse topological sort index (as superscript of the node letter) (see Section 2.2) and with their $\text{suff}(x)$ values (as subscript of the node letter) (see Section 3.1). Arcs are labeled with their PPH increments.

## 2. VITERBI ALGORITHM

Viterbi Algorithm computes the likelihood that a given HMM generates a given string of symbols by using *Dynamic Programming* [1, 9].

As illustrated in **Figure** 2, each path in a lexicon-HMM joining the start state to a terminal state corresponds uniquely to a word of the lexicon and reciprocally.

Let $a_{ij}$ be the transition probability from state $i$ to state $j$ in the lexicon-HMM, and let $b_j(o)$ be the emission probability of symbol $o$ in state $j$. Finally, let $\delta_t(j)$ be the maximum log probability (score of the most likely sequence of hidden states) of the HMM model being in state $j$ after generating the sequence of symbols $(o_1,...,o_t)$.

The time series $\delta_t(j)$ satisfy the following recurrence relation, the *Standard Viterbi Decoder Equation* (SVDE)

$$\delta_t(j) = \max_{i \in \text{pred}(j)} \{\delta_{t-1}(i) + \log(a_{ij})\} + \log(b_j(o_t))$$

If for each state $j$, we record its maximizing predecessor, we can easily determine the sequence of states that is the most likely to generate the sequence of observed symbols $(o_1,...,o_T)$, by first identifying which state $\hat{j}_T$ maximizes $\delta_T(j)$, then tracing back (*backtracking*) the rest of the most likely sequence of states with the sequence of maximizing predecessors starting from state $j_T$.

### 2.1. Basic token passing implementation (arbitrary sort)

As Young et al [15] pointed out, at any time step, only a single value $\delta_t(j)$, and the corresponding best partial path $\hat{p}_t(j)$, needs to be stored in state $j$. This approach is best viewed as a systolic propagation of tokens over a network.

In a token passing implementation of VA, each node $j$ holds a single token $\tau_t(j)$, containing $\delta_t(j)$ and some information $\hat{p}_t(j)$ to represent the (partial) optimal path (called *Path History*), from the root node, via the sequence of maximizing predecessors.

**Pseudo-code** 1. Basic token passing implementation of VA
```
Initialize the token of the root.
  for t = 1:T
    for each j ∈ (j₁, j₂,..., j_N) // arbitrary order
      Compute τ_t(j) using the SVDE on τ_{t-1}(i).
    endfor
  endfor
```

In practice, the values of $\tau_t(j)$ and $\tau_{t-1}(j)$ are stored in two different flip-flop pointed arrays. Although the running time of this implementation of VA is optimal, some memory space is wasted.

### 2.2. Memory space optimized token passing implementation (reverse topological sort)

If we ignore the loops of the states, the HMM that we consider are acyclic. Recall that a sequence $(j_1,...,j_N)$ of states is compatible with the *Topological Sort* [12] if the sequence is an indexing of the states such that if $a < b$ then $j_a$ is not a descendant of $j_b$. The superscripts of the nodes of the toy DAWG node-automaton in **Figure** 2 are derived from such a sequence. A simple change in the order which the tokens are passed halves the memory requirement of VA:

**Pseudo-code** 2. Space optimized token passing implementation of VA[1]

```
// PRE: (j₁, j₂,..., j_N) is a sequence compatible with
// the topological sort. This sequence is computed
// only once for a given HMM.
Initialize the token of the root.
  for t = 1:T
    // Scan states in reverse topological sort.
    for each j ∈ (j_N, j_{N-1},..., j₁)
      Compute τ_t(j) using the SVDE on τ_{t-1}(i).
    endfor
  endfor
```

Because we visit the states in reverse topological order, the same memory variable can be used to store $\tau_t(j)$ and $\tau_{t-1}(j)$. When $\tau_{t-1}(j)$ is overwritten by $\tau_t(j)$, the value of $\tau_{t-1}(j)$ is no longer needed. Whereas, if the states were scanned in the order $(j_1, j_2,..., j_N)$, then $\tau(j)$ would require two distinct memory variables; one for time $t$ and one for time $t-1$. It is a special case of the scheduling problem with precedence constraints in DAG [12]. Moreover, if node data are stored following reverse topological sort, VA update steps can be done by simply incrementing a memory address pointer (thereby achieving optimal memory access speed).

---

[1] A more detailed, and improved, form of loop internal pseudo-code is presented later for the $n$-best case and can apply as well to 1-best. The computational complexity stays the same.

### 2.3. $n$-best Token Passing Viterbi Algorithm

It is often desirable in practice to determine not just the best path (sequence of states or nodes depending the resolution) that maximizes the total score, but the $n$-best different paths [8]. In the context of this paper, different paths mean different words. Indeed, considering hypotheses other than the one corresponding to the best path increases the chances of finding the correct word. In many applications, some information not used in the HMM recognizer, such as a more precise grammar or a language model or other contextual clues, is used to re-rank the candidate words (improving the recognition rate).

Thanks to Bellman principle of optimality [1], it is sufficient to keep the list of the best $n$ tokens at each state in order to determine the $n$-best paths. This list is a sorted list of $n$ tokens $\tau_t(j,n) \equiv (\tau_t(j,1),...,\tau_t(j,n))$ where $\tau_t(j,1)$ is the best token.

**Pseudo-code 3.** Naïve merging of $n$-best solutions in token passing VA

```
// Update of the n-best token τ_t(j,n) of state j
// at time t.
Initialize τ_t(j,n) to void.
  for each i ∈ pred(j) // Scan the predecessors of
j.
    for k = 1:n // if any
      On the token τ_{t-1}(i,k) use update equations
      δ_t(j) = (δ_{t-1}(i,k) + log(a_{ij})) + log(b_j(o_t))
      p̂_t(j) = p̂_{t-1}(i,k) + s_j
      to build a candidate token τ_t(j) to be merged
      in sorted list τ_t(j,n). Keep only, if any,
      n tokens with different p̂_t(j)
    endfor
  endfor
```

In the above naïve implementation of the $n$-best VA the inner loop, including merging, is executed systematically $n \times |\text{pred}(j)|$ times. So the time complexity is $n$ times the 1-best complexity plus the complexity of merging operations.

The order of execution of the loops does matter. If the order of the next pseudo-code is chosen, a simplified merging operation can occur:

**Pseudo-code 4.** Improved merging of $n$-best solutions in token passing VA

```
// Improved update of the n-best token τ_t(j,n)
// of state j at time t.
Initialize τ_t(j,n) to void.
  for k = 1:n
    for each i ∈ pred(j)
      From token τ_{t-1}(i,k) calculate
```
$$\delta_t(j) = \delta_{t-1}(i,k) + \log(a_{ij})$$
```
      Test if  δ_t(j)  has to be merged in
```
$(\tau_t(j,k),...,\tau_t(j,n))$
```
      If "merged", update the token τ_t(j) with
```
$\hat{p}_t(j) = \hat{p}_{t-1}(i,k) + s_j$ and if needed, delete the
```
      worst token with same  p̂_t(j)
    endfor
```
$\delta_t(j,k) = \delta_t(j,k) + \log(b_j(o_t))$ in $\tau_t(j,k)$
```
  endfor
```

Some update operations can be conditional or extracted from the most internal loop leading to significantly more efficient computations:
- Incrementation of best partial path is restricted to merged tokens.
- Final incrementation is only, and usefully, done on the $n$-best tokens for the step.
- Due to the principle of optimality [1], a simplified merging operation occurs "from the $k^{th}$ element to the end of the list" only if needed.

### 2.4. Time and space complexities

Let $N$ be the total number of states of the HMM, and $T$ be the length of the input sequence of symbols. It is easy to see that the time complexity (theoretical worst case) of the tabular (basic) implementation of VA is $O(N^2 T)$. The factor $N^2$ is the product of the number of states ($N$) times the maximum in-degree ($N$). However, the maximum in-degree of a lexicon-HMM derived from real languages is in practice independent from $N$, and much smaller than $N$. The relevant factor is $p = \frac{1}{N} \sum |\text{pred}(j)|$. For example, an actual 130 K words French lexicon [6] gives:

| trie | $N = 297701$ | $p \equiv 1$ |
|---|---|---|
| DAWG | $N = 17908$ | $p = 5.22$ |

**Table 1.** Lexicons at http://webia.lip6.fr/~lifchitz/FLCVA.

So the average time complexity is $O(NT)$, as for the space one, as summarized below:

|  | time (worst) | time (average) | space | best path |
|---|---|---|---|---|
| **tabular** | $O(N^2 T)$ | $O(NT)$ | $O(NT)$ | backtracking |
| **token passing** | $O(N^2 T)$ | $O(NT)$ | $O(N)$ | path history |

**Table 2.** Time and space complexities of the different implementations of VA.

## 3. AN OPTIMAL TOKEN TAGGING SCHEME FOR VITERBI ALGORITHM

The order in which the full paths (from root to sink) of an automaton are completed in a *Depth First Search* (DFS) [12] provides a canonical indexing of the full paths of the automaton [11]. We call this index the *Perfect Path History* (PPH) as it is a *Minimal Perfect Hashing* [2] and is perfectly suited to the management of path history. This PPH index is naturally extended to partial paths (from root to internal node), and constitutes an optimal coding scheme for the paths followed by the tokens. In the rest of this section, we show how to compute the PPH.

### 3.1. An optimal coding for the paths of the node-automaton

Consider a DAWG of a lexicon of $W$ words. For a node $x$ of the automaton, let $\text{suff}(x)$ denotes the number of paths (suffixes) from this node to the sink. In particular $\text{suff}(root) = W$ and $\text{suff}(sink) = 1$, as the empty path is a valid path.

We have the recursive definition:
$$\text{suff}(x) = \begin{cases} 1 & \text{if } x \text{ is the } sink \\ \sum_i \text{suff}(\text{succ}(x,i)) & \text{otherwise} \end{cases}$$

There is a canonical injection from the set of partial paths to the set of full paths. To each partial path $p$, we associate $\bar{p}$ the full path with the smallest $\text{PPH}(p)$ that is an extension of $p$. Therefore, we can extend the PPH() function defined originally on the full paths to the partial paths by defining $\text{PPH}(p)$ as $\text{PPH}(\bar{p})$.

The value of $\text{PPH}(p)$ is by construction in the range $[0, W-1]$ and:
$$\text{PPH}(root) \equiv 0$$
as the extension of the root node partial path to a full path is always the first full path in the DFS order.

Let $p(x)$ be a partial path from the root node to $x$. It is easy to see that, if
$$\Delta\text{PPH}(x, \text{succ}(x,i)) = \sum_{j<i} \text{suff}(\text{succ}(x,j))$$

| Word | PPH |
|---|---|
| ab | 0 |
| ba | 1 |
| bb | 2 |
| bc | 4 |
| bcd | 3 |
| c | 5 |

**Table 3.** The six words (listed in alphabetical order) of the toy automaton **Figure** 2 and their PPH.

then
$$\text{PPH}(p(x) + \text{succ}(x,i)) = \text{PPH}(p(x)) + \Delta\text{PPH}(x, \text{succ}(x,i))$$

### 3.2. Important properties of the PPH coding scheme

The PPH coding is a bijection between paths (beginning in root node) in DAWG and integer values in the range $[0, W-1]$.

$\text{PPH}(p(x))$ is the path history information carried by the token when it is at node $x$ and has followed the partial path $p(x)$. The above formula is used to update this token PPH when it is passed from node $x$ to node $\text{succ}(x,i)$.

The PPH increment $\Delta\text{PPH}(x, \text{succ}(x,i))$ can either be computed dynamically (minimizing memory space requirement), or can be cached in each arc $x \to \text{succ}(x,i)$ (maximizing speed) as in **Figure** 2. Thanks to its recursive definition, $\text{suff}(x)$ can be computed with a recursive DFS. The complexity of this traversal is linear in the number of nodes. This computation needs to be done only once for a given DAG, as an initialization process.

The PPH exhibits interesting properties that makes it a perfectly suited coding system for the partial paths. In particular, the PPH can be computed with only local information. The size of the PPH values in bits is optimal as it takes its value in the range $[0, W-1]$. Given the PPH value of a token at a given node, it is easy to trace the path followed by the token from the root: complexity is linear in the number of nodes of this path.

## 4. CONCLUSION

In this paper, we have outlined a novel approach for lexically constrained HMM based recognition systems. Our proposed approach, the Fast Lexically Constrained Viterbi Algorithm (FLCVA), combines several optimization techniques:
- Compact DAWG in place of traditional tries with simultaneous gain in memory space and running time (typically a ratio 15-20 for a 100 K words lexicon) because of the large reduction of the overall number of HMM states to consider.
- Non deterministic node-automata in place of classical arc-automata for better compacity.

- Reverse topological sort of nodes that halves memory requirement.
- Enhanced $n$-best algorithm.
- Optimal token tagging scheme for path history management (PPH).

A prototype program was written in Java (non optimized code). The running times on a desktop PC (Pentium P4 3 GHz 512 MB RAM) on the previously mentioned 130 K words lexicon (Table 1) demonstrates a dramatic speed-up:

| trie | 202 seconds |
|------|-------------|
| DAWG | 11 seconds  |

**Table 4**. CPU times for a 130Kwords French lexicon, 1-best, on the same machine, for trie (classical approach) and DAWG (FLCVA).

**Figures**

**Pseudo-codes**

**Tables**